\documentclass[letterpaper]{article}
\usepackage{aaai25}
\usepackage{times}
\usepackage{helvet}
\usepackage{courier}
\usepackage[hyphens]{url}
\usepackage{graphicx}
\usepackage{algorithm}
\usepackage{algorithmic}
\usepackage{subcaption}
\usepackage{booktabs}
\usepackage{multirow}
\usepackage{amsfonts}
\usepackage{amsmath}
\usepackage{dsfont}
\usepackage{bibentry}

\usepackage{natbib}
\usepackage{caption}
\usepackage{bibentry}
\usepackage{newfloat}
\usepackage{listings}

\DeclareCaptionStyle{ruled}{labelfont=normalfont,labelsep=colon,strut=off}
\floatstyle{ruled}
\newfloat{listing}{tb}{lst}{}
\floatname{listing}{Listing}

\title{Distilling Knowledge from Heterogeneous Architectures \\for Semantic Segmentation}

\author{
    Yanglin Huang\textsuperscript{\rm 1},
    Kai Hu\textsuperscript{\rm 1, }\thanks{Corresponding author.},
    Yuan Zhang\textsuperscript{\rm 1},
    Zhineng Chen\textsuperscript{\rm 2},
    Xieping Gao\textsuperscript{\rm 3}
}
\affiliations{
    \textsuperscript{\rm 1}Key Laboratory of Intelligent Computing and Information Processing of Ministry of Education, Xiangtan University \\
    \textsuperscript{\rm 2}School of Computer Science, Fudan University \\
    \textsuperscript{\rm 3}Key Laboratory for Artificial Intelligence and International Communication, Hunan Normal University\\
    ylhuang@smail.xtu.edu.cn, \{kaihu, yuanz\}@xtu.edu.cn, zhinchen@fudan.edu.cn, xpgao@hunnu.edu.cn
}

\begin{document}

\maketitle

\begin{abstract}
Current knowledge distillation (KD) methods for semantic segmentation focus on guiding the student to imitate the teacher's knowledge within homogeneous architectures. However, these methods overlook the diverse knowledge contained in architectures with different inductive biases, which is crucial for enabling the student to acquire a more precise and comprehensive understanding of the data during distillation. To this end, we propose for the first time a generic knowledge distillation method for semantic segmentation from a heterogeneous perspective, named \textit{HeteroAKD}. Due to the substantial disparities between heterogeneous architectures, such as CNN and Transformer, directly transferring cross-architecture knowledge presents significant challenges. To eliminate the influence of architecture-specific information, the intermediate features of both the teacher and student are skillfully projected into an aligned logits space. Furthermore, to utilize diverse knowledge from heterogeneous architectures and deliver customized knowledge required by the student, a teacher-student knowledge mixing mechanism (KMM) and a teacher-student knowledge evaluation mechanism (KEM) are introduced. These mechanisms are performed by assessing the reliability and its discrepancy between heterogeneous teacher-student knowledge. Extensive experiments conducted on three main-stream benchmarks using various teacher-student pairs demonstrate that our \textit{HeteroAKD} outperforms state-of-the-art KD methods in facilitating distillation between heterogeneous architectures.
\end{abstract}

\begin{figure}[t]
\centering
\includegraphics[width=0.98\linewidth]{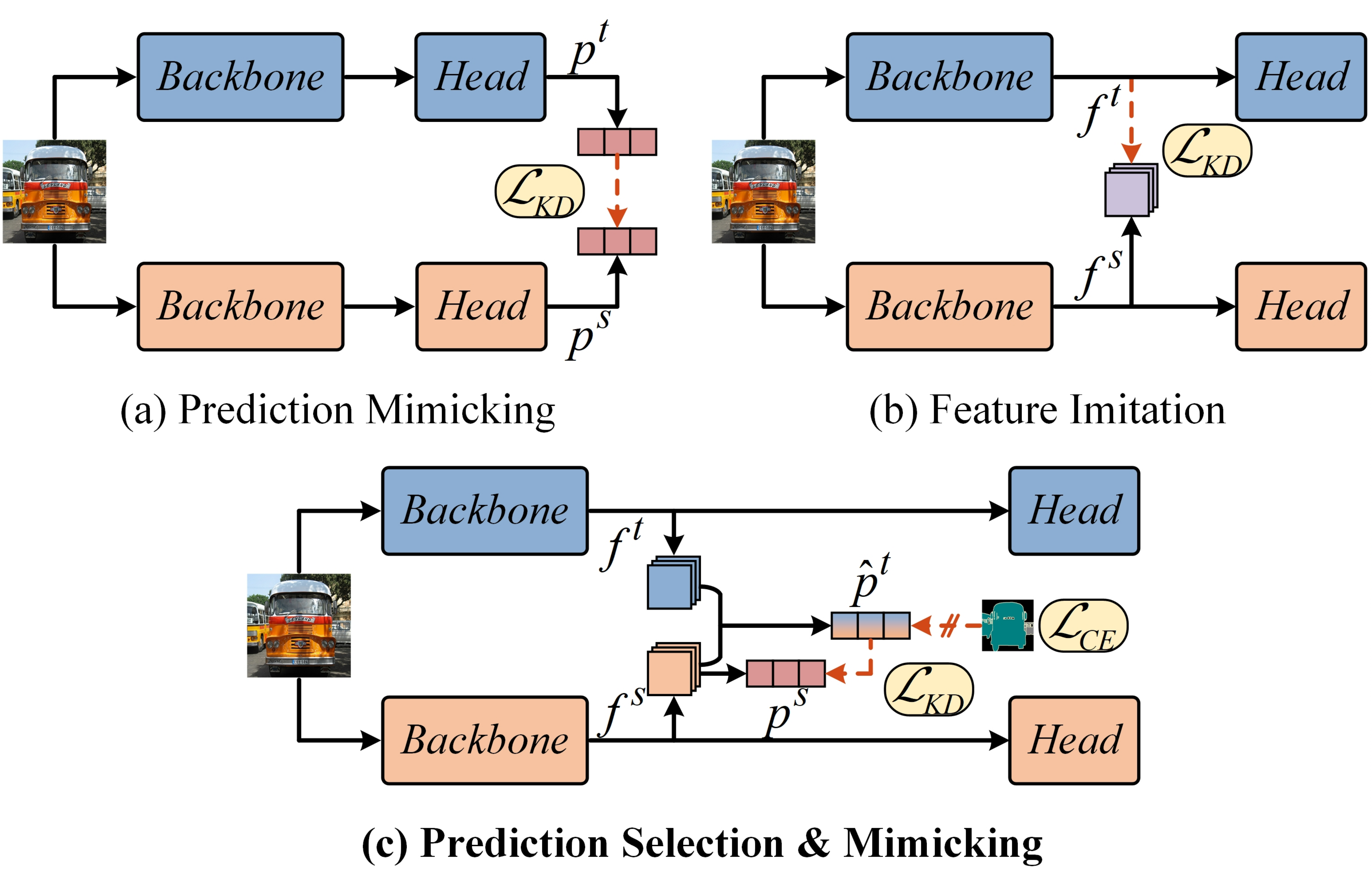}
\caption{Comparison of the vanilla KD methods ((a) and (b)) with our \textit{HeteroAKD} (c).}
\label{fig1}
\end{figure}

\section{Introduction}
Knowledge Distillation (KD), as a model compression technique, has been extensively researched in the field of semantic segmentation and has achieved remarkable progress~\cite{liu2019s,he2019k,wang2020i,shu2021c,yang2022c,fan2023a}. 
According to the distillation position of the segmenters, existing KD methods can be roughly classified into two categories: logits-based and feature-based. Logits-based methods (See Figure~\ref{fig1}a) follow the idea proposed in~\cite{hinton2015d}, which forces the student to mimic the prediction distribution of the teacher to acquire more accurate knowledge. Differently, feature-based methods (See Figure~\ref{fig1}b), inspired by~\cite{romero2015f}, extend the form of taught knowledge from the prediction distribution to the feature representation of the model. It aims to enforce the feature consistency between the teacher-student pair.

Currently, both logits-based~\cite{shu2021c,baek2022d} and feature-based~\cite{he2019k,liu2022m} KD approaches for semantic segmentation focus on knowledge transfer between teacher-student pairs in homogeneous architectures, while the distillation of heterogeneous architectures has not been explored. However, it is crucial to distill knowledge from heterogeneous architectures in practical scenarios. In general, architectures with different inductive biases tend to focus on distinct patterns, enabling them to understand the data from various perspectives to attain diverse knowledge~\cite{ren2022c}. Therefore, gaining diverse knowledge from heterogeneous architectures enables students to achieve a more precise and comprehensive understanding of the data during distillation. For example, when distilling the student model on the ADE20K~\cite{zhou2019s} dataset, as our experiments will demonstrate, transferring knowledge from DeepLabV3-ResNet-101 to SegFormer-Mix Transformer-B1 (our \textit{HeteroAKD} $\Delta$mIoU: +3.66\%) can easily surpass the performance increment achieved by transferring knowledge to DeepLabV3-ResNet-18 (Af-DCD~\cite{fan2023a} recorded $\Delta$mIoU: +2.30\%). Distilling knowledge from heterogeneous architectures thus provides another viable solution. Moreover, the continuous emergence of new architectures~\cite{chen2022c,gu2023m} brings deeper understanding of the data, allowing researchers to enhance their own models using pre-trained teachers of different architectures.

\begin{figure}[t]
\centering
\includegraphics[width=1.0\linewidth]{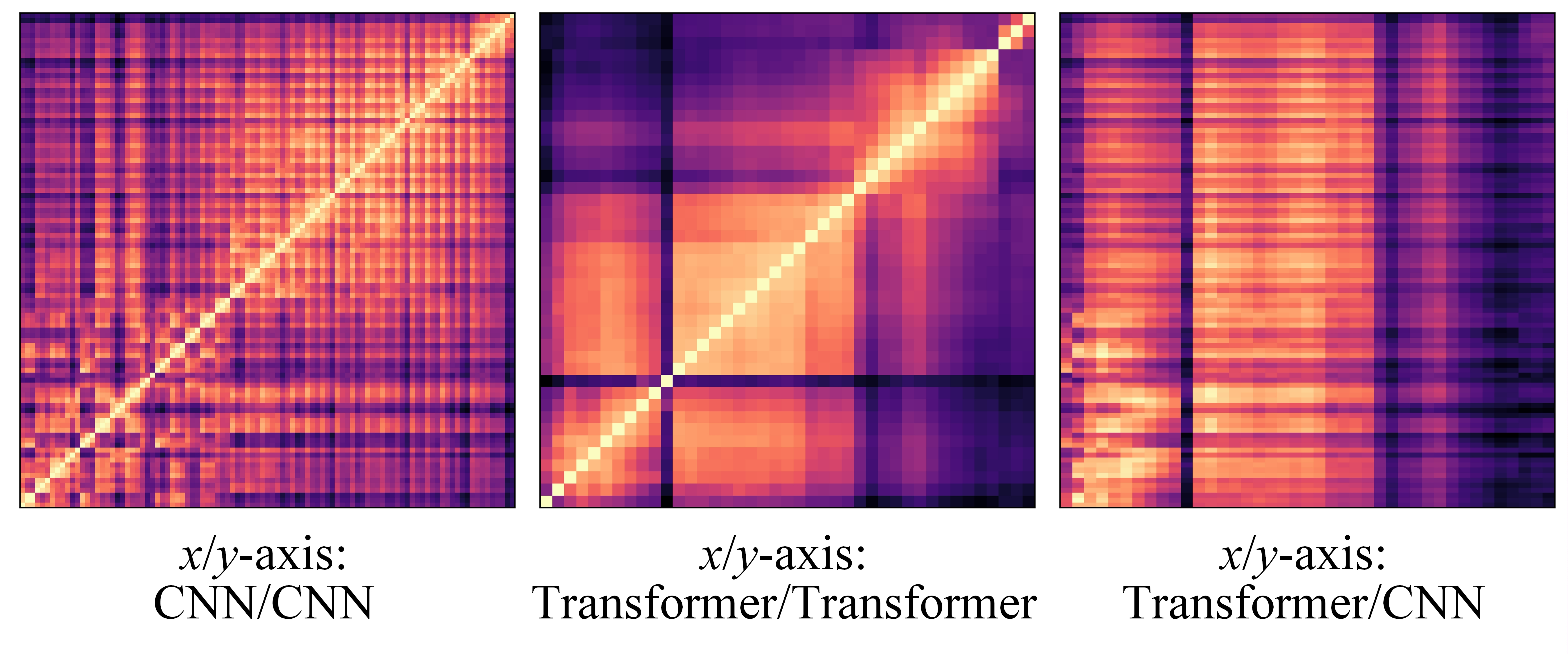}
\caption{Similarity heatmap of intermediate features measured by centered kernel alignment (CKA). We compare features from ResNet-101 (CNN) and Mix Transformer-B4 (Transformer). Best viewed with zoom in.}
\label{fig2}
\end{figure}

Due to the substantial disparities between heterogeneous architectures, directly transferring knowledge from the teacher to the student presents significant challenges. This prompts us to consider: \textit{how can a student effectively extract knowledge while retaining its own expertise when faced with a heterogeneous teacher?} Through an in-depth investigation of two types of main-stream architectures in KD, \textit{i.e.}, CNN and Transformer, we argue that existing KD approaches face two key challenges: \textbf{(i)} the substantial disparities in the features learned by teachers and students with different inductive biases, as illustrated in Figure~\ref{fig2}; \textbf{(ii)} the uncritical imitation may lead students to acquire erroneous knowledge which is caused by the fact that prediction made by teachers are not invariably superior to those made by students (See Figure~\ref{fig3}).

To this end, we propose for the first time a generic \textbf{Hetero}geneous \textbf{A}rchitecture \textbf{K}nowledge \textbf{D}istillation framework for semantic segmentation, named \textit{HeteroAKD} (See Figure~\ref{fig1}c). 
To tackle the first challenge, instead of using any fancy tricks to bridge the intermediate feature gap between heterogeneous teacher-student pairs, we transfer the feature representations into the aligned logits space, which contains less architecture-specific information. By matching the output of the student's intermediate features with that of the teacher's in logits space, the student is constrained to approximate the teacher's performance. 
This manner to knowledge transfer in logits space circumvents directly imposing constraints on the students' intermediate features, thereby allowing the student more flexibility in learning intermediate feature representations that are conducive to downstream tasks~\cite{zheng2023l}.

To address the second challenge, we utilize human knowledge (\textit{i.e.}, labels) to serve as the “textbook”, guiding the process of knowledge transfer for students. Inspired by human educational practices~\cite{midgley2014g}, we propose a teacher-student knowledge mixing mechanism (KMM) and a teacher-student knowledge evaluation mechanism (KEM), to utilize diverse knowledge from heterogeneous architectures and deliver customized knowledge desired by the student. Specifically, prior to targeted instruction, the KMM assesses the reliability of knowledge by calculating the loss between intermediate feature outputs of both teacher and student against labels. This assessment guides the dynamic generation of more precise teacher-student hybrid knowledge, which incorporates contributions from both the teacher and student.
Due to varying levels of student mastery of different knowledge at different times~\cite{yang2024l}, directly imitating teacher-student hybrid knowledge may not be an optimal choice. The KEM further utilizes the knowledge reliability discrepancy between teacher and student to evaluate the relative importance of knowledge, which can deliver the customized knowledge according to the student's ability. As the learning progresses, the KEM progressively guides the student to master more difficult knowledge to increase the upper performance limit.

In summary, our main contributions are listed as follows:
\begin{itemize}
    \item We propose a novel \textit{HeteroAKD} framework, which transfers heterogeneous architecture knowledge in the logits space, to eliminate the influence of architecture-specific information. To the best of our knowledge, this is the first generic knowledge distillation method for semantic segmentation explored from a heterogeneous perspective.
    \item We propose a teacher-student knowledge mixing mechanism and a teacher-student knowledge evaluation mechanism based on human knowledge guidance to utilize diverse knowledge from heterogeneous architectures and deliver customized knowledge desired by the student.
    \item Extensive experiments on three main-stream benchmarks demonstrate the superiority of our \textit{HeteroAKD} in facilitating distillation between heterogeneous architectures.
\end{itemize}

\section{Related Work}
\noindent{\textbf{Knowledge Distillation.}} 
KD is an effective method for transferring valuable knowledge from a complex teacher model to a simpler student model. Currently, KD methods can be broadly categorized into logits-based and feature-based approaches according to the distillation position. Logits-based KD methods~\cite{hinton2015d,zhou2021r} required the student model to replicate the class probability distribution of the teacher model. Feature-based KD methods~\cite{romero2015f,chen2021c,chen2021d,hao2022l} transferred detailed feature activation from the teacher model to supervise the learning process of the student model. As models with diverse inductive biases offer a more comprehensive depiction of the data, recent methods have begun to incorporate distillation techniques based on heterogeneous architectures, leading to promising performance across various tasks, such as classification~\cite{ren2022c, hao2023o}, face recognition~\cite{zhao2023c} and monocular depth estimation~\cite{zheng2024p}.
\newline

\noindent{\textbf{Knowledge Distillation in Semantic Segmentation.}}
Since semantic segmentation is an intensive predictive task, direct application of KD methods designed for other tasks may not yield satisfactory results. Thus, specific KD methods have been proposed for semantic segmentation. For example, SKD~\cite{liu2019s} directly aligned the similarity between teacher-student pairs at the pixel-wise level, while CWD~\cite{shu2021c} transferred meaningful knowledge by simply minimizing the channel-wise pixel distribution between teacher-student pairs. In addition, Af-DCD~\cite{fan2023a} proposed a contrastive distillation learning paradigm to utilize feature partitions across both channel and spatial dimensions for knowledge transfer. Furthermore, some methods explore the inherent knowledge among different samples. Among them, IFVD~\cite{wang2020i} forced the student to mimic teacher intra-class relations by assessing distances with prototypes from different classes. 
Similarly, CIRKD~\cite{yang2022c} built pixel dependencies across global samples to transfer structured relations knowledge. Despite achieving remarkable performance in existing distillation methods for semantic segmentation, they assume that the student and teacher architectures are homogeneous. However, when the architectures are heterogeneous, these methods may fail due to significant variability between the student and teacher. C2VKD~\cite{zheng2023d} attempts to learn a compact Transformer-based model from a cumbersome yet high-performance CNN-based model, but this approach is still limited to transforming knowledge in a single mode. Therefore, how to distill knowledge from any heterogeneous architectures for semantic segmentation remains an open problem.

\section{Methodology}

\subsection{Preliminary}
\subsubsection{Notations of Knowledge Distillation.} Logits and features are the most common used types of knowledge in KD. A naive logits-based method is to train the student to mimic the class probability distribution of each pixel of the teacher, which can be defined as:
\begin{equation}
\centering
    \mathcal L_{kd}=\frac{1}{H{\times}W}\sum\limits_{h=1}^H {\sum\limits_{w=1}^W {KL({\sigma({\frac{{\mathbf{Z}_{h,w}^s}}{\tau}})\| {\sigma({\frac{{\mathbf{Z}_{h,w}^t}}{\tau}})}})}},
\label{Eq1}
\end{equation}
\noindent where $\sigma({\mathbf{Z}_{h,w}^s}/{\tau})$ and $\sigma({\mathbf{Z}_{h,w}^t}/{\tau})$ denote the soft class probabilities of the student and teacher models on the ($h$, $w$)-th pixel, respectively. $KL(\cdot)$ represents the Kullback-Leibler divergence function. $\tau$ is a temperature parameter.

Different from the logits-based method, the feature-based method encourages the student to mimic the more fine-grained teacher feature activation. The formulation can be expressed as:
\begin{equation}
\centering
    \mathcal L_{fd}=\frac{1}{H{\times}W}\sum\limits_{h=1}^H {\sum\limits_{w=1}^W  {{({\mathbf{F}_{h,w}^t - \psi({\mathbf{F}_{h,w}^s})})^2}}},
\label{Eq2}
\end{equation}
\noindent where $\mathbf{F}_{h,w}^t$ and $\mathbf{F}_{h,w}^s$ denote the ($h$, $w$)-th pixel in features produced from the teacher and student models, respectively. $\psi(\cdot)$ is a feature projector that maps student model features to match the dimension of teacher model features.

\subsection{Analysis of Knowledge Distillation for Heterogeneous Architectures}
\label{sec:3.2}
To explore the impacts of the intrinsic differences of heterogeneous architectures (\textit{i.e.}, CNN and Transformer) on knowledge distillation for semantic segmentation, we provide an analysis of logits-based and feature-based methods of knowledge distillation.

\begin{figure}[t]
\centering
\includegraphics[width=1.0\linewidth]{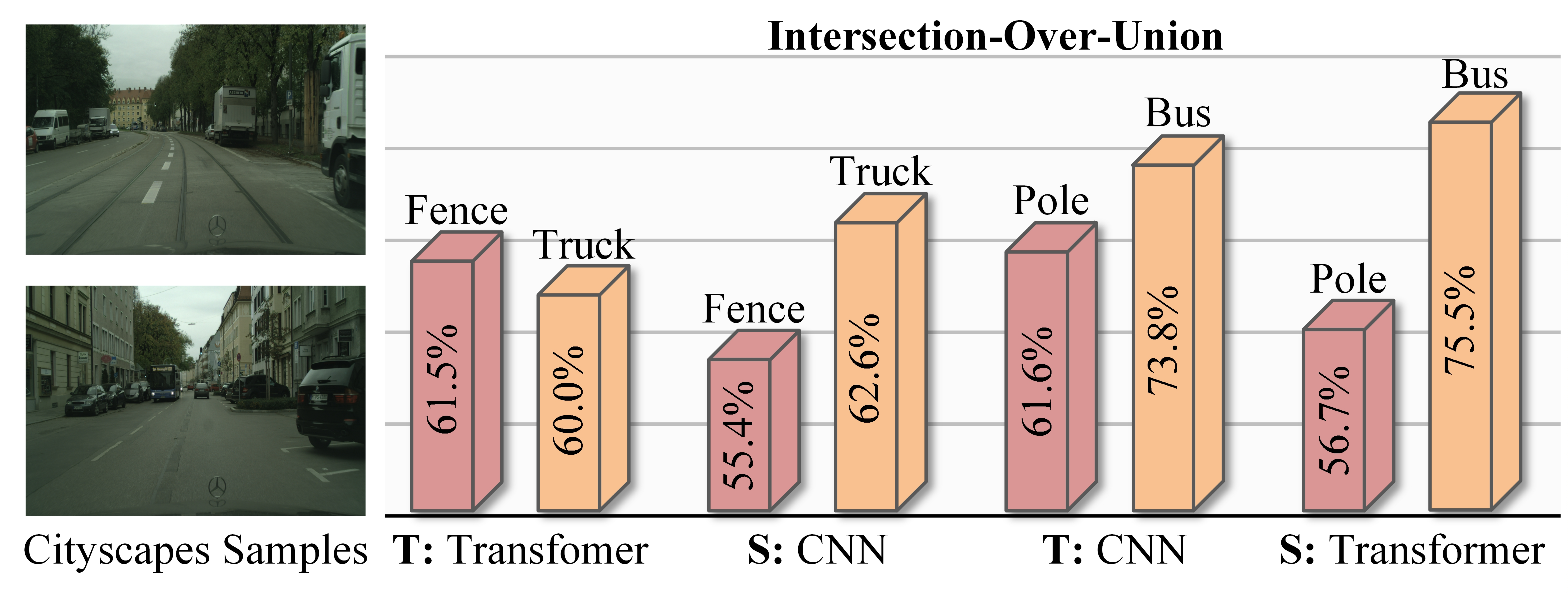}
\caption{Analysis of IoU metrics for class probabilities predicted by CNN-based and Transformer-based architectures. We choose the first pair of teacher-student models for each mode in Table~\ref{tab:cityscape(b)} for our analysis.}
\label{fig3}
\end{figure}

\begin{figure*}[t]
\centering
\includegraphics[width=1.0\linewidth]{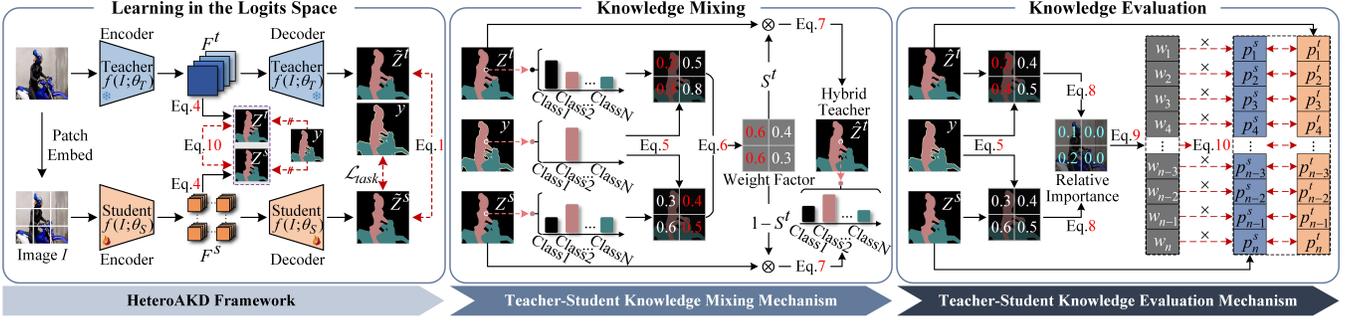}
\caption{An overview of the \textit{HeteroAKD} framework. Here, we take the “CNN$\rightarrow$Transformer” mode as an example.}
\label{fig4}
\end{figure*}

\subsubsection{Centered Kernel Alignment Analysis.} Inspired by~\cite{hao2023o}, we employ minibatch centered kernel alignment (CKA)~\cite{kornblith2019s,nguyen2021d} to compare the feature representations extracted by heterogeneous architectures in semantic segmentation. Suppose $\mathbf{X}_{i}\in{\mathbb{R}^{n\times {d_1}}}$ and $\mathbf{Y}_{i}\in{\mathbb{R}^{n\times {d_2}}}$ are features of the $i$-th minibatch of $n$ samples extracted by CNN-based and Transformer-based models, with $d_1$ and $d_2$ neurons respectively. Let $\mathbf{K}_i = \mathbf{X}_i\mathbf{X}^\mathsf{T}_i$ and $\mathbf{L}_i = \mathbf{Y}_i\mathbf{Y}^\mathsf{T}_i$ denote the Gram matrices for the two feature representations (which reflects the similarities between a pair of samples according to feature representations), CKA can be computed as:
\begin{equation}
\centering
\mathrm{CKA}=\frac{\frac{1}{k}\sum\limits_{i=1}^k{\mathrm{HSIC}(\mathbf{K}_i,\mathbf{L}_i)}}{{\sqrt{\frac{1}{k}\sum\limits_{i=1}^k{\mathrm{HSIC}(\mathbf{K}_i,\mathbf{K}_i)}}}{\sqrt{\frac{1}{k}\sum\limits_{i=1}^k{\mathrm{HSIC}(\mathbf{L}_i,\mathbf{L}_i)}}}},
\label{Eq3}
\end{equation}
\noindent where $k$ denotes the number of minibatch. $\mathrm{HSIC}$ is the Hilbert-Schmidt independence criterion~\cite{gretton2007a}. In our implementation, we use an unbiased estimator of $\mathrm{HSIC}$ as proposed in~\cite{song2012f}.

From Figure~\ref{fig2}, we can observe that homogeneous architectures prefer to learn similar feature representations at layers of similar positions, whereas heterogeneous architectures only achieve similar feature representations at shallow layers. 
Existing feature-based distillation methods directly project teacher and student features to the same dimension, which is not a universal solution for aligning feature representations of heterogeneous architectures. 
\textit{How to project features into a space that is unaffected by architecture-specific information is a key aspect in designing heterogeneous distillation methods.}

\subsubsection{Class Probabilities Analysis.} 
Heterogeneous architectures exhibit significant differences in their inner features and output paradigms, which often leads to different class distributions~\cite{huang2024c}. An intuitive idea is that a complex teacher is not always superior to a simple student. Instead, we believe that architectures with different inductive biases tend to learn more precise knowledge on particular patterns. To this end, we analyze the IoU metrics for class probabilities predicted by heterogeneous architectures, as shown in Figure~\ref{fig3}. We can observe that teachers are inferior to the corresponding heterogeneous students in specific classes (\textit{e.g.}, truck and bus). This indicates that heterogeneous architectures produces inconsistent understanding of knowledge from different perspectives, even if they are learning from the same dataset. 
Existing logits-based methods naively mimic the teacher's class probability distribution, which may lead to students acquiring erroneous knowledge. \textit{How to utilize the diverse knowledge from heterogeneous architectures and deliver the knowledge required by the student is another key aspect in designing heterogeneous distillation methods.}

\begin{table*}[!t]
    \small
    \centering
    \begin{subtable}[t]{0.495\linewidth}
        \centering
        \begin{tabular}{l|c|c|c}
            \toprule[1pt]
            \textbf{Method} & \textbf{Params} & \textbf{FLOPs} & \textbf{Val mIoU} \\
            \hline
            \hline
            \multicolumn{4}{l}{\textit{Mode: Transformer$\rightarrow$CNN}} \\
            \hline
            \hline
            T: DeepLabV3-MiT-B4 & 63.5M & 980.1G & 75.89 \\
            \hline
            S: DeepLabV3-Res18 & \multirow{7}{*}{13.6M} & \multirow{7}{*}{572.0G} & 74.53 \\
            +SKD & & & 73.55 \\
            +IFVD & & & 74.54 \\
            +CWD & & & 73.39 \\
            +CIRKD & & & 73.88 \\
            +Af-DCD & & & \underline{75.64} \\
            +HeteroAKD (Ours) & & & \textbf{76.35} \\
            \hline
            S: DeepLabV3-MBV2 & \multirow{7}{*}{4.1M} & \multirow{7}{*}{164.9G} & 73.92 \\
            +SKD & & & 71.50 \\
            +IFVD & & & 73.25 \\
            +CWD & & & 71.59 \\
            +CIRKD & & & 73.20 \\
            +Af-DCD & & & \underline{73.69} \\
            +HeteroAKD (Ours) & & & \textbf{74.91} \\
            \hline
            \hline
            \multicolumn{4}{l}{\textit{Mode: CNN$\rightarrow$Transformer}} \\
            \hline
            \hline
            T: DeepLabV3-Res101 & 61.1M & 2371.7G & 78.34 \\
            \hline
            S: DeepLabV3-MiT-B1 & \multirow{7}{*}{15.8M} & \multirow{7}{*}{275.9G} & 70.91 \\
            +SKD & & & 72.04 \\
            +IFVD & & & 72.43 \\
            +CWD & & & 73.31 \\
            +CIRKD & & & 72.87 \\
            +Af-DCD & & & \underline{73.81} \\
            +HeteroAKD (Ours) & & & \textbf{74.28} \\
            \hline
            S: DeepLabV3-PVT-B1 & \multirow{7}{*}{16.1M} & \multirow{7}{*}{293.9G} & 71.90 \\
            +SKD & & & 72.74 \\
            +IFVD & & & 73.32 \\
            +CWD & & & \underline{73.94} \\
            +CIRKD & & & 73.69 \\
            +Af-DCD & & & 73.81 \\
            +HeteroAKD (Ours) & & & \textbf{74.65} \\
            \bottomrule[1pt]
        \end{tabular}
        \caption{The same segmentation head with different backbone architectures}
        \label{tab:cityscape(a)}
    \end{subtable}
    \begin{subtable}[t]{0.495\linewidth}
    \centering
        \begin{tabular}{l|c|c|c}
            \toprule[1pt]
            \textbf{Method} & \textbf{Params} & \textbf{FLOPs} & \textbf{Val mIoU} \\
            \hline
            \hline
            \multicolumn{4}{l}{\textit{Mode: Transformer$\rightarrow$CNN}} \\
            \hline
            \hline
            T: SegFormer-MiT-B4 & 64.1M & 1230.1G & 78.80 \\
            \hline
            S: DeepLabV3-Res18 & \multirow{7}{*}{13.6M} & \multirow{7}{*}{572.0G} & 74.53 \\
            +SKD & & & 74.28 \\
            +IFVD & & & 75.16 \\
            +CWD & & & 73.53 \\
            +CIRKD & & & 74.68 \\
            +Af-DCD & & & \underline{75.46} \\
            +HeteroAKD (Ours) & & & \textbf{76.42} \\
            \hline
            S: PSPNet-Res18 & \multirow{7}{*}{12.9M} & \multirow{7}{*}{507.4G} & 73.19 \\
            +SKD & & & 71.19 \\
            +IFVD & & & 72.94 \\
            +CWD & & & \underline{73.74} \\
            +CIRKD & & & 72.60 \\
            +Af-DCD & & & 71.97 \\
            +HeteroAKD (Ours) & & & \textbf{74.26} \\
            \hline
            \hline
            \multicolumn{4}{l}{\textit{Mode: CNN$\rightarrow$Transformer}} \\
            \hline
            \hline
            T: DeepLabV3-Res101 & 61.1M & 2371.7G & 78.34 \\
            \hline
            S: SegFormer-MiT-B1 & \multirow{7}{*}{13.7M} & \multirow{7}{*}{240.3G} & 74.91 \\
            +SKD & & & 70.68 \\
            +IFVD & & & 73.73 \\
            +CWD & & & 74.80 \\
            +CIRKD & & & 74.25 \\
            +Af-DCD & & & \underline{75.20} \\
            +HeteroAKD (Ours) & & & \textbf{76.34} \\
            \hline
            S: PSPNet-MiT-B1 & \multirow{7}{*}{15.1M} & \multirow{7}{*}{247.0G} & 71.29 \\
            +SKD & & & 67.06 \\
            +IFVD & & & 73.29 \\
            +CWD & & & \underline{73.41} \\
            +CIRKD & & & 72.68 \\
            +Af-DCD & & & 72.99 \\
            +HeteroAKD (Ours) & & & \textbf{74.25} \\
            \bottomrule[1pt]
        \end{tabular}
        \caption{Different segmentation heads with the same backbone architecture}
        \label{tab:cityscape(b)}
    \end{subtable}
    \caption{Comparison with state-of-the-art distillation methods on Cityscapes validation set. ‘T’ and ‘S’ denote the teacher and student, respectively. Params and FLOPs are measured according to CIRKD~\cite{yang2022c}. The \textbf{best}/\underline{second best} results are marked in bold/underline.}
    \label{tab:cityscape}
\end{table*}

\subsection{Proposed Heterogeneous Architecture Knowledge Distillation}
An overview of the proposed \textit{HeteroAKD} framework is illustrated in Figure~\ref{fig4}. Our aim is to train a compact student model $f({I}; \theta_{S})$ by transferring diverse knowledge from a heterogeneous teacher model $f({I}; \theta_{T})$. This student model $f({I}; \theta_{S})$ possesses a more precise and comprehensive understanding of the data, enabling accurate assignment of a pixel-wise label $y_{h,w} \in 1,...,C$ to each pixel $p_{h,w}$ in image $i \in I$. Next, we will elaborate in detail on the key components that drive our framework.

\subsubsection{Learning in the Logits Space.} Given the input images $I$, we can obtain the intermediate feature representations ($\mathbf{F}^t\in{\mathbb{R}^{H_{1}\times W_{1} \times d_1}}$ and $\mathbf{F}^s\in{\mathbb{R}^{H_{2}\times W_{2} \times d_2}}$) from the heterogeneous teacher model $f({I}; \theta_{T})$ and student model $f({I}; \theta_{S})$. As analyzed in Section~\ref{sec:3.2}, directly aligning feature representations $\mathbf{F}^t$ and $\mathbf{F}^s$ is extremely challenging. To this end, we propose to project the intermediate features of the teacher $\mathbf{F}^t$ and student $\mathbf{F}^s$ into the logits space, thereby obtaining their respective categorical logit maps, designated as $\mathbf{Z}^t\in{\mathbb{R}^{H\times W \times C}}$ and $\mathbf{Z}^s\in{\mathbb{R}^{H\times W \times C}}$, respectively. Here, $H$ and $W$ are the height and width of image $i\in I$, and $C$ is the number of classes. $\mathbf{Z}^t$ and $\mathbf{Z}^s$ eliminate redundant architecture-specific information, and thus provide an ideal form of transferring knowledge from heterogeneous architectures~\cite{hao2023o}. Moreover, performing knowledge distillation in the logits space circumvents directly imposing constraints on student's intermediate features $\mathbf{F}^s$, thereby allowing the student model $f({I}; \theta_{S})$ more flexibility in learning feature representations that are conducive for downstream tasks~\cite{zheng2023l}. This process can be formulated as:
\begin{equation}
\centering
\mathbf{Z}^t = \mathcal{G}_{proj}(\mathbf{F}^t), \quad \mathbf{Z}^s = \mathcal{G}_{proj}(\mathbf{F}^s), 
\label{Eq4}
\end{equation}
\noindent where $\mathcal{G}_{proj}(\cdot)$ denotes a feature projector that is composed of 1$\times$1 convolutional layer with BN and ReLU.

\begin{table*}[!t]
    \small
    \centering
    \begin{subtable}[t]{0.495\linewidth}
        \centering
        \begin{tabular}{l|c|c|c}
            \toprule[1pt]
            \textbf{Method} & \textbf{Params} & \textbf{FLOPs} & \textbf{Val mIoU} \\
            \hline
            \hline
            \multicolumn{4}{l}{\textit{Mode: Transformer$\rightarrow$CNN}} \\
            \hline
            \hline
            T: SegFormer-MiT-B4 & 64.1M & 485.8G & 80.27 \\
            \hline
            S: DeepLabV3-Res18 & \multirow{7}{*}{13.6M} & \multirow{7}{*}{305.0G} & 74.53 \\
            +SKD & & & 74.08 \\
            +IFVD & & & 73.75 \\
            +CWD & & & 71.43 \\
            +CIRKD & & & \underline{74.87} \\
            +Af-DCD & & & 74.18 \\
            +HeteroAKD (Ours) & & & \textbf{75.44} \\
            \hline
            \hline
            \multicolumn{4}{l}{\textit{Mode: CNN$\rightarrow$Transformer}} \\
            \hline
            \hline
            T: DeepLabV3-Res101 & 61.1M & 1294.6G & 78.82 \\
            \hline
            S: SegFormer-MiT-B1 & \multirow{7}{*}{13.7M} & \multirow{7}{*}{89.0G} & 75.66 \\
            +SKD & & & 72.70 \\
            +IFVD & & & 74.70  \\
            +CWD & & & 74.79 \\
            +CIRKD & & & 75.23 \\
            +Af-DCD & & & \underline{75.73} \\
            +HeteroAKD (Ours) & & & \textbf{76.11} \\
            \bottomrule[1pt]
        \end{tabular}
        \caption{Pascal VOC}
        \label{tab:Pascal VOC}
    \end{subtable}
    \begin{subtable}[t]{0.495\linewidth}
        \centering
        \begin{tabular}{l|c|c|c}
            \toprule[1pt]
            \textbf{Method} & \textbf{Params} & \textbf{FLOPs} & \textbf{Val mIoU} \\
            \hline
            \hline
            \multicolumn{4}{l}{\textit{Mode: Transformer$\rightarrow$CNN}} \\
            \hline
            \hline
            T: SegFormer-MiT-B4 & 64.1M & 485.8G & 46.20 \\
            \hline
            S: DeepLabV3-Res18 & \multirow{7}{*}{13.6M} & \multirow{7}{*}{305.0G} & 33.70 \\
            +SKD & & & 34.38 \\
            +IFVD & & & 34.54 \\
            +CWD & & & 33.09 \\
            +CIRKD & & & \underline{35.05} \\
            +Af-DCD & & & 34.68 \\
            +HeteroAKD (Ours) & & & \textbf{35.73} \\
            \hline
            \hline
            \multicolumn{4}{l}{\textit{Mode: CNN$\rightarrow$Transformer}} \\
            \hline
            \hline
            T: DeepLabV3-Res101 & 61.1M & 1294.6G & 42.47 \\
            \hline
            S: SegFormer-MiT-B1 & \multirow{7}{*}{13.7M} & \multirow{7}{*}{89.0G} & 35.18 \\
            +SKD & & & 33.57 \\
            +IFVD & & & 34.95  \\
            +CWD & & & 33.74 \\
            +CIRKD & & & 34.71 \\
            +Af-DCD & & & \underline{36.74} \\
            +HeteroAKD (Ours) & & & \textbf{38.84} \\
            \bottomrule[1pt]
        \end{tabular}
        \caption{ADE20K}
        \label{tab:ADE20K}
    \end{subtable}
    \caption{Comparison with state-of-the-art distillation methods on Pascal VOC and ADE20K validation sets. Params and FLOPs are measured according to CIRKD~\cite{yang2022c}. The \textbf{best}/\underline{second best} results are marked in bold/underline.}
    \label{tab:other dataset}
\end{table*}

\subsubsection{Teacher-Student Knowledge Mixing Mechanism.}
As analyzed in Section~\ref{sec:3.2}, the teacher may not outperform the student on a particular pattern. Our objective is to generate a teacher-student hybrid knowledge, which has a more precise and comprehensive understanding of the data. To this end, we treat labels as “textbook” that contain reliable knowledge generated by human intelligence. 
Thereby, the knowledge reliability of each pixel can be obtained by calculating the cross-entropy between the pixel-wise label $\mathbf{y}_{h,w}$ and class probability distribution $\sigma(\mathbf{Z}_{h,w})$ as:

\begin{equation}
\begin{aligned}
\centering
\mathcal{H}(\mathbf{Z}_{h,w|c})= &-(\mathbf{y}_{h,w}\log(\sigma(\mathbf{Z}_{h,w|c})) \\
                              &+(1-\mathbf{y}_{h,w})\log(1-\sigma(\mathbf{Z}_{h,w|c}))),
\label{Eq5}
\end{aligned}
\end{equation}
\noindent where $\mathbf{Z}_{h,w|c}$ denotes the categorical logit map at the position of $(h,w)$ of the $c$-th channel. $\sigma(\cdot)$ is the sigmoid function. 
A lower cross-entropy value indicates a greater degree of similarity between the probability distribution of $\sigma(\mathbf{Z}_{h,w|c})$ and $\mathbf{y}_{h,w}$. Accordingly, we argue that a lower cross-entropy value reflects a higher degree of knowledge reliability. 
Following the established criteria for knowledge reliability, we perform a preference selection of teacher's and student's knowledge, which can be formulated as follows:
\begin{equation}
\centering
\mathbf{S}_{h,w|c}^t=1-\frac{\mathcal{H}(\mathbf{Z}_{h,w|c}^t)}{\mathcal{H}(\mathbf{Z}_{h,w|c}^t)+\mathcal{H}(\mathbf{Z}_{h,w|c}^s)},
\label{Eq6}
\end{equation}
\noindent where $\mathbf{S}_{h,w|c}^t$ records the weight factor of pixels from the teacher, while $(1-\mathbf{S}_{h,w|c}^t)$ denotes the weight factor of pixels from the student at the corresponding position. 
According to the obtained weight factor, we can select a more accurate hybrid knowledge $\mathbf{\hat{Z}}_{h,w|c}^t$ from both the teacher and student. It can be formulated as follows:
\begin{equation}
\centering
\mathbf{\hat{Z}}_{h,w|c}^t=\mathbf{S}_{h,w|c}^t\odot\mathbf{Z}_{h,w|c}^t+(1-\mathbf{S}_{h,w|c}^t)\odot\mathbf{Z}_{h,w|c}^s,
\label{Eq7}
\end{equation}
\noindent where $\odot$ denotes Hadamard product. Notably, it is difficult to obtain valuable information by directly utilizing $\mathbf{Z}_{h,w|c}^s$ generated by a naive student trained from scratch. In fact, such an approach may even compromise the accuracy of the hybrid knowledge $\mathbf{\hat{Z}}_{h,w|c}^t$. Therefore, we warm up the student model $f({I}; \theta_{S})$ under the full supervision of labels $\mathbf{y}$ before distillation.

\subsubsection{Teacher-Student Knowledge Evaluation Mechanism.}
During distillation, an important pixel is one that the student has not yet fully grasped, but which can be acquired through learning from the teacher. 
To this end, we propose to evaluate the relative importance of pixels through the discrepancy between the hybrid teacher and student knowledge reliability, which can be utilized as a guidance to provide customized knowledge to the student. It is formulated as:
\begin{equation}
\centering
\Delta\mathcal{H}(\mathbf{Z}_{h,w|c}) = \mathds{1}_{+} \times(\mathcal{H}(\mathbf{Z}_{h,w|c}^s) - \mathcal{H}(\mathbf{\hat{Z}}_{h,w|c}^t)),
\label{Eq8}
\end{equation}
\noindent where $\mathds{1}_{+}$ is an indicator function which returns $1$ if $\mathcal{H}(\mathbf{Z}_{h,w|c}^s)>\mathcal{H}(\mathbf{\hat{Z}}_{h,w|c}^t)$ else $0$. $\Delta\mathcal{H}(\mathbf{Z}_{h,w|c})$ denotes the relative importance of the pixel at the position $(h,w)$ of the $c$-th channel. We further transform the relative importance of pixels into weight values by:

\begin{equation}
\centering
\mathbf{W}_{:,:|c} =
\begin{cases}
  \frac{\exp(\mathcal{H}(\mathbf{Z}_{:,:|c}^s)+\Delta\mathcal{H}(\mathbf{Z}_{:,:|c}))}{\sum\limits_{i=1}^C\exp(\mathcal{H}(\mathbf{Z}_{:,:|i}^s)+\Delta\mathcal{H}(\mathbf{Z}_{:,:|i}))}, & \Delta\mathcal{H}(\mathbf{Z}_{:,:|c}) > 0 \\
 \frac{\exp(\mathcal{H}(\mathbf{Z}_{:,:|c}^s))}{\sum\limits_{i=1}^C\exp(\mathcal{H}(\mathbf{Z}_{:,:|i}^s)+\Delta\mathcal{H}(\mathbf{Z}_{:,:|i}))}, & \Delta\mathcal{H}(\mathbf{Z}_{:,:|c}) \leq 0 \\
\end{cases}
\label{Eq9}
\end{equation}
\noindent where $\mathbf{W}_{:,:|c}$ denotes the weight matrix of $c$-th category. According to $\mathbf{W}_{:,:|c}$, we reweight the original distillation loss (Eq.~\ref{Eq1}) to enhance the important information desired by the student as:
\begin{equation}
\centering
\mathcal L_{hakd}=-\frac{1}{C}{\sum\limits_{c=1}^C{\sigma(\frac{\mathbf{\hat{Z}}_{:,:|c}^t}{\tau})\log(\sigma(\frac{\mathbf{Z}_{:,:|c}^s}{\tau}))\times\mathbf{W}_{:,:|c}}},
\label{Eq10}
\end{equation}
\noindent where $\sigma({\mathbf{\hat{Z}}_{:,:|c}^t}/{\tau})$ and $\sigma({\mathbf{Z}_{:,:|c}^s}/{\tau})$ denote the soft class probabilities of the hybrid teacher and student on the $c$-th category. $\tau$ is a temperature parameter.

\subsection{Optimization Objective}
The overall loss for optimization can be formulated as the weighted sum of the task loss $\mathcal L_{task}$, class probability KD loss $\mathcal L_{kd}$ (Eq.~\ref{Eq1}), and heterogeneous architecture KD loss $\mathcal L_{hakd}$ (Eq.~\ref{Eq10}), written as:
\begin{equation}
\centering
    \mathcal L_{total}=\mathcal L_{task} + \lambda_{1}\mathcal L_{kd} + \lambda_{2}\mathcal L_{hakd},
\end{equation}
\noindent where $\mathcal L_{task}$ is the cross-entropy loss for semantic segmentation task. $\lambda_{1}$ and $\lambda_{2}$ are weight factors used to balance the relationship between losses. In distillation losses, the projection heads used for teacher-student pairwise dimension matching are composed of 1$\times$1 convolutional layer with BN and ReLU. They are discarded at the inference phase without introducing extra costs. Notably, the features from the last layer of the backbone architecture are used for distillation in our $\mathcal L_{hakd}$.

\section{Experiments}

\subsection{Experimental Setups}

\subsubsection{Datasets.} Our experiments are conducted on three popular semantic segmentation datasets, including Cityscapes~\cite{cordts2016t}, Pascal VOC~\cite{everingham2016t} and ADE20K~\cite{zhou2019s}.

\begin{figure}[t]
\centering
\includegraphics[width=1.0\linewidth]{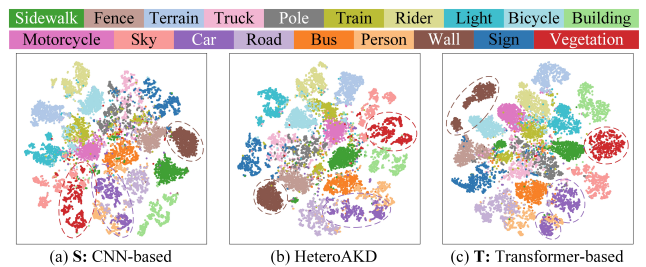}
\caption{T-SNE visualization of learned feature embeddings (\textit{i.e.}, SegFormer-MiT-B4$\rightarrow$DeepLabV3-Res18) on the Cityscapes dataset. We outline some classes with dash circles in their colors for a clearer view.}
\label{fig5}
\end{figure}

\subsubsection{Implementation Details.} Following the previous methods~\cite{liu2019s,yang2022c,fan2023a}, we adopt DeepLabV3~\cite{chen2018e}, PSPNet~\cite{zhao2017p} and SegFormer~\cite{xie2021s} for segmentation heads, ResNet-101 (Res101)~\cite{he2016d} and Mix Transformer-B4 (MiT-B4)~\cite{xie2021s} for teacher backbone architectures, ResNet-18 (Res18), MobileNetV2 (MBV2)~\cite{sandler2018m}, Mix Transformer-B1 (MiT-B1) and Pyramid Vision Transformer v2-B1 (PVT-B1)~\cite{wang2022p} for student backbone architectures and group various teacher-student pairs. We compare the proposed \textit{HeteroAKD} with state-of-the-art (SOTA) knowledge distillation methods for semantic segmentation: SKD~\cite{liu2019s}, IFVD~\cite{wang2020i}, CWD~\cite{shu2021c}, CIRKD~\cite{yang2022c} and Af-DCD~\cite{fan2023a}. We re-implemented all methods on both CIRKD codebase~\cite{yang2022c} and Af-DCD codebase~\cite{fan2023a}. For crop size during the training phase, we use 512$\times$1024, 512$\times$512 and 512$\times$512 for Cityscapes, Pascal VOC and ADE20K, respectively.

\subsection{Comparison with State-of-the-Art Methods}

\subsubsection{Results on Cityscapes.} Table~\ref{tab:cityscape} presents the quantitative results of four backbone architectures and three segmentation heads on Cityscapes dataset. Our \textit{HeteroAKD} consistently outperforms the baseline across all backbone architectures, with the maximum mIoU and average mIoU margin by 3.37\% and 2.04\%, respectively. Notably, in some cases (\textit{e.g.}, DeepLabV3-MiT-B4$\rightarrow$DeepLabV3-Res18), the student's performance after distillation is superior to that of the teacher by 0.46\%. This indicates that students are not simply imitating their teachers to learn knowledge. In contrast, existing SOTA KD methods are heavily influenced by the challenges analyzed in Section~\ref{sec:3.2}, making it difficult to benefit from heterogeneous teachers.

As shown in Figure~\ref{fig5}, we further analyze the feature embeddings learned by our \textit{HeteroAKD} using T-SNE visualization. The visual results indicate that our \textit{HeteroAKD} maintains its own advantageous feature embeddings while pushing students to imitate the teacher's feature embeddings. This facilitates students to achieve better intra-class compactness and inter-class separability, thus improving segmentation performance.

\begin{table}[!t]
    \small
    \centering
    \begin{tabular}{l|c|c}
        \toprule[1pt]
        \textbf{Method} & \textbf{mIoU (\%)} & \textbf{$\Delta$mIoU (\%)} \\
        \hline
        \multicolumn{3}{l}{\textit{Mode: Transformer$\rightarrow$CNN}} \\
        \hline
        \textit{Baseline} & 74.53 & n/a  \\
        +$\mathcal L_{kd}$ & 75.67 & +1.14 \\
        +$\mathcal L_{hakd}$ & 76.03 & +1.50 \\
        +$\mathcal L_{kd}$ + $\mathcal L_{hakd}$ & 76.42 & +1.89 \\
        +$\mathcal L_{kd}$ + $\mathcal L_{hakd}$ \textit{w/o} \textit{KMM} & 76.19 & +1.66 \\
        +$\mathcal L_{kd}$ + $\mathcal L_{hakd}$ \textit{w/o} \textit{KEM} & 75.82 & +1.29 \\
        \hline
        \multicolumn{3}{l}{\textit{Mode: CNN$\rightarrow$Transformer}} \\
        \hline
        \textit{Baseline} & 74.91 & n/a  \\
        +$\mathcal L_{kd}$ & 75.64 & +0.73 \\
        +$\mathcal L_{hakd}$ & 75.56 & +0.65 \\
        +$\mathcal L_{kd}$ + $\mathcal L_{hakd}$ & 76.34 & +1.43 \\
        +$\mathcal L_{kd}$ + $\mathcal L_{hakd}$ \textit{w/o} \textit{KMM} & 75.87 & +0.96 \\
        +$\mathcal L_{kd}$ + $\mathcal L_{hakd}$ \textit{w/o} \textit{KEM} & 75.96 & +1.05 \\
        \bottomrule[1pt]
    \end{tabular}
    \caption{Ablation studies of loss terms and key components on Cityscapes validation set. The results are obtained using the first teacher-student pair for each mode in Table~\ref{tab:cityscape(b)}.}
    \label{tab:loss}
\end{table}

\subsubsection{Results on Two Other Datasets.} In Table~\ref{tab:other dataset}, we compare the proposed \textit{HeteroAKD} with the existing SOTA KD methods on PASCAL VOC and ADE20K datasets to validate the generalization of our method in solving different semantic segmentation tasks. According to the results shown in Table~\ref{tab:Pascal VOC}-\ref{tab:ADE20K}, our \textit{HeteroAKD} consistently achieves the best performance in different heterogeneous distillation modes for different datasets. Compared to the SOTA KD methods, our model gains the maximal mIoU and average mIoU margin by 2.10\% and 0.93\%, respectively. The results demonstrate that our \textit{HeteroAKD} is effective in facilitating knowledge distillation between heterogeneous teacher-student pairs in different semantic segmentation tasks.

\subsection{Ablation Studies}

\subsubsection{Ablation Study on Different Loss Terms.} We analyze the contribution of each distillation loss. From the results shown in Table~\ref{tab:loss}, we can get following observations: (i) Compared to \textit{Baseline}, the introduction of either $\mathcal{L}_{kd}$ (0.94\% average mIoU gain) or $\mathcal{L}_{hakd}$ (1.08\% average mIoU gain) individually improves the performance of both knowledge transfer modes. (ii) The baseline continues to demonstrate improvement (1.66\% average mIoU gain), with the combined contribution of both losses $\mathcal{L}_{kd}+\mathcal{L}_{hakd}$. This indicates that simultaneous learning from teacher intermediate features and output logits is beneficial for improving the student performance.

\subsubsection{Ablation Study on Key Components.} We verify the validity of the proposed KMM and KEM in $\mathcal{L}_{hakd}$. As shown in Table~\ref{tab:loss}, we can see that removing either KMM (0.35\% average mIoU reduction) or KEM (0.49\% average mIoU reduction) brings a significant negative impact on performance. This illustrates that both components are instrumental in distilling knowledge from heterogeneous architectures.

\subsubsection{Ablation Studies on Hyper-parameters.} 
\begin{figure}[t]
\centering
\includegraphics[width=1.0\linewidth]{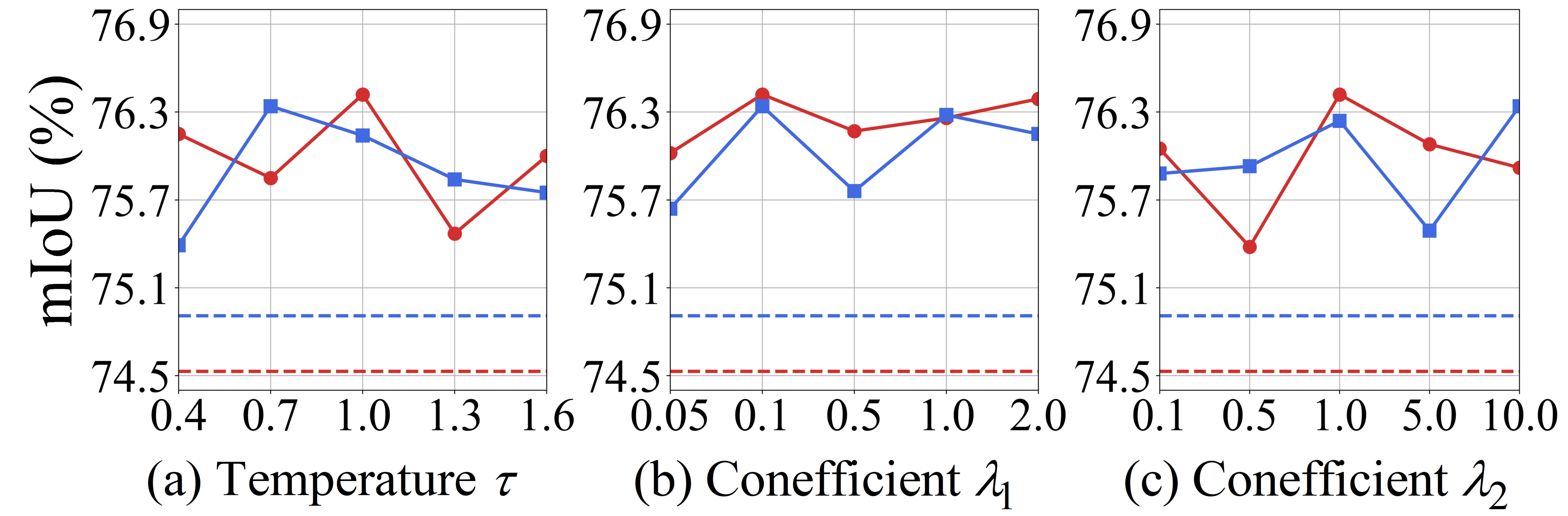}
\caption{Ablation studies of (a) temperature $\tau$, (b) $\mathcal L_{kd}$ coefficient $\lambda_{1}$ and (c) $\mathcal L_{hakd}$ coefficient $\lambda_{2}$ on Cityscapes validation set. Red and blue lines indicate “Transformer$\rightarrow$CNN” and “CNN$\rightarrow$Transformer” modes, respectively.}
\label{fig6}
\end{figure}

We investigate the impact of different hyper-parameter settings. As illustrated in Figure~\ref{fig6}, our method consistently enables students to benefit from heterogeneous teachers, with the minimum mIoU gain of 0.48\%. Different hyper-parameter settings have different impacts on distillation efficiency, this difference in optimal hyper-parameters can be attributed to the varying strengths of the teacher and student.

\section{Conclusion}
In this paper, we propose a generic knowledge distillation framework for semantic segmentation from a heterogeneous perspective, named \textit{HeteroAKD}. 
Compared to previous methods, our \textit{HeteroAKD} can help students learn more diverse knowledge from the heterogeneous teacher.
Extensive experiments on three main-stream benchmarks demonstrate the superiority of our \textit{HeteroAKD} framework in facilitating distillation between heterogeneous architectures. While our method makes significant progress in facilitating distillation between heterogeneous architectures, it is worth noting that in certain cases, the efficiency of knowledge distillation from a heterogeneous teacher may be lower than that achieved by a homogeneous teacher.

\section*{Acknowledgements}

This work was supported in part by the National Natural Science Foundation of China under Grants 62272404 and 62372170, in part by the Natural Science Foundation of Hunan Province of China under Grant 2023JJ40638, and in part by the Research Foundation of Education Department of Hunan Province of China under Grant 23A0146.

\bibliography{aaai25}

\newpage

\appendix

\section{Experimental Setups}

\subsection{Datasets}
\noindent \textbf{Cityscapes}~\cite{cordts2016t} is an urban scene parsing dataset that contains 2,975/500/1,525 finely annotated images for \texttt{train/val/test}. The segmentation performance is reported on 19 classes. 

\noindent \textbf{Pascal VOC}~\cite{everingham2016t} is a visual object segmentation dataset, which contains 20 foreground classes and 1 background class. We employ the augmented dataset with extra annotations provided by~\cite{hariharan2011s} resulting in 10,582/1,449 images for \texttt{train/val}. 

\noindent \textbf{ADE20K}~\cite{zhou2019s} is a challenging semantic segmentation dataset, which contains 20,210/2,000/3,352 images for \texttt{train/val/test}, with totally 150 classes.

\subsection{Evaluation Metrics}
Following the general setting~\cite{yang2022c,fan2023a}, we use mean Intersection-over-Union (mIoU) to measure the segmentation performance of all methods. Furthermore, we adopt the network parameters (Params) and the sum of floating point operations (FLOPs) on a fixed input size to show the model size and complexity.

\subsection{Training Details}
Following the general settings~\cite{yang2022c,fan2023a}, random flipping and scaling in the range of [0.5, 2] are employed to augment the data. The ‘poly’ learning rate policy~\cite{chen2017r} is utilized, where the learning rate is decayed by $(1-\frac{iter}{total\_iter})^{0.9}$. 
The total training iterations are set to 80K for Cityscapes, and 40K for Pascal VOC and ADE20K.
We use a batch size of 8 for Cityscapes, and a batch size of 16 for Pascal VOC and ADE20K.
For crop size during the training phase, we use 512$\times$1024, 512$\times$512 and 512$\times$512 for Cityscapes, Pascal VOC, and ADE20K, respectively. 
We employ different optimizers for training the student models based on their encoder architecture. Specifically, the CNN-based students are trained using SGD optimizer with the weight decay of 0.0001 and the initial learning rate of 0.01; while the Transformer-based students are trained using the AdamW optimizer with the weight decay of 0.0001 and the initial learning rate of 0.00006. All experiments are trained on 4 A100 GPUs.

\subsection{Evaluation Details}
We evaluate the segmentation performance under a single scale setting over the original image size following the general protocol~\cite{shu2021c}.

\subsection{CKA Analysis Settings}
In the Centered Kernel Alignment (CKA) analysis, we evaluate the features extracted from two pre-trained backbone architectures: ResNet-101~\cite{he2016d} and Mix Transformer-B4~\cite{xie2021s}. For the analysis, 500 samples are selected from the Cityscapes dataset, and the minibatch size is set to 10.

\section{More Experimental Results}

\subsection{Results on Homogeneous Architectures}
To further demonstrate the effectiveness of our \textit{HeteroAKD} on homogeneous architectures, we conduct experiments on CNN-based and Transformer-based architectures. As shown in Table~\ref{tab:homogeneous architectures}, we can observe that our \textit{HeteroAKD} achieves promising performance, which brings 1.61\% average mIoU gain to baseline student model. 
Moreover, we observe that the improvement in performance ($\Delta$mIoU: +0.62\%) achieved from a homogeneous teacher (DeepLabV3-Res101) exceeds that derived from a heterogeneous teacher (SegFormer-MiT-B4) when training a DeepLabV3-Res18 student. 
The experimental results not only validate the efficacy of our \textit{HeteroAKD} in distilling knowledge from homogeneous teachers but also suggest that, in certain scenarios, it may be more challenging for the student to distill knowledge from a heterogeneous teacher compared to a homogeneous one. 
As such, there remains an open question regarding how to improve the efficiency of knowledge transfer from heterogeneous teachers to students.

\begin{table}[!t]
    \small
    \centering
    \begin{tabular}{l|c|c|c}
        \toprule[1pt]
        \textbf{Method} & \textbf{Params} & \textbf{FLOPs} & \textbf{Val mIoU} \\
        \hline
        \hline
        \multicolumn{4}{l}{\textit{Mode: Transformer$\rightarrow$Transformer}} \\
        \hline
        \hline
        T: SegFormer-MiT-B4 & 64.1M & 1230.1G & 78.80 \\
        \hline
        S: SegFormer-MiT-B1 & \multirow{3}{*}{13.7M} & \multirow{3}{*}{240.3G} & 74.91 \\
        +Af-DCD & & & 75.37 \\
        +HeteroAKD (Ours) & & & \textbf{75.62} \\
        \hline
        \hline
        \multicolumn{4}{l}{\textit{Mode: CNN$\rightarrow$CNN}} \\
        \hline
        \hline
        T: DeepLabV3-Res101 & 61.1M & 2371.7G & 78.34 \\
        \hline
        S: DeepLabV3-Res18 & \multirow{3}{*}{13.6M} & \multirow{3}{*}{572.0G} & 74.53 \\
        +Af-DCD & & & 76.41 \\
        +HeteroAKD (Ours) & & & \textbf{77.04} \\
        \bottomrule[1pt]
    \end{tabular}
    \caption{Comparison with state-of-the-art distillation methods on Cityscapes validation set.}
    \label{tab:homogeneous architectures}
\end{table}

\begin{figure*}[t]
\centering
\includegraphics[width=0.83\linewidth]{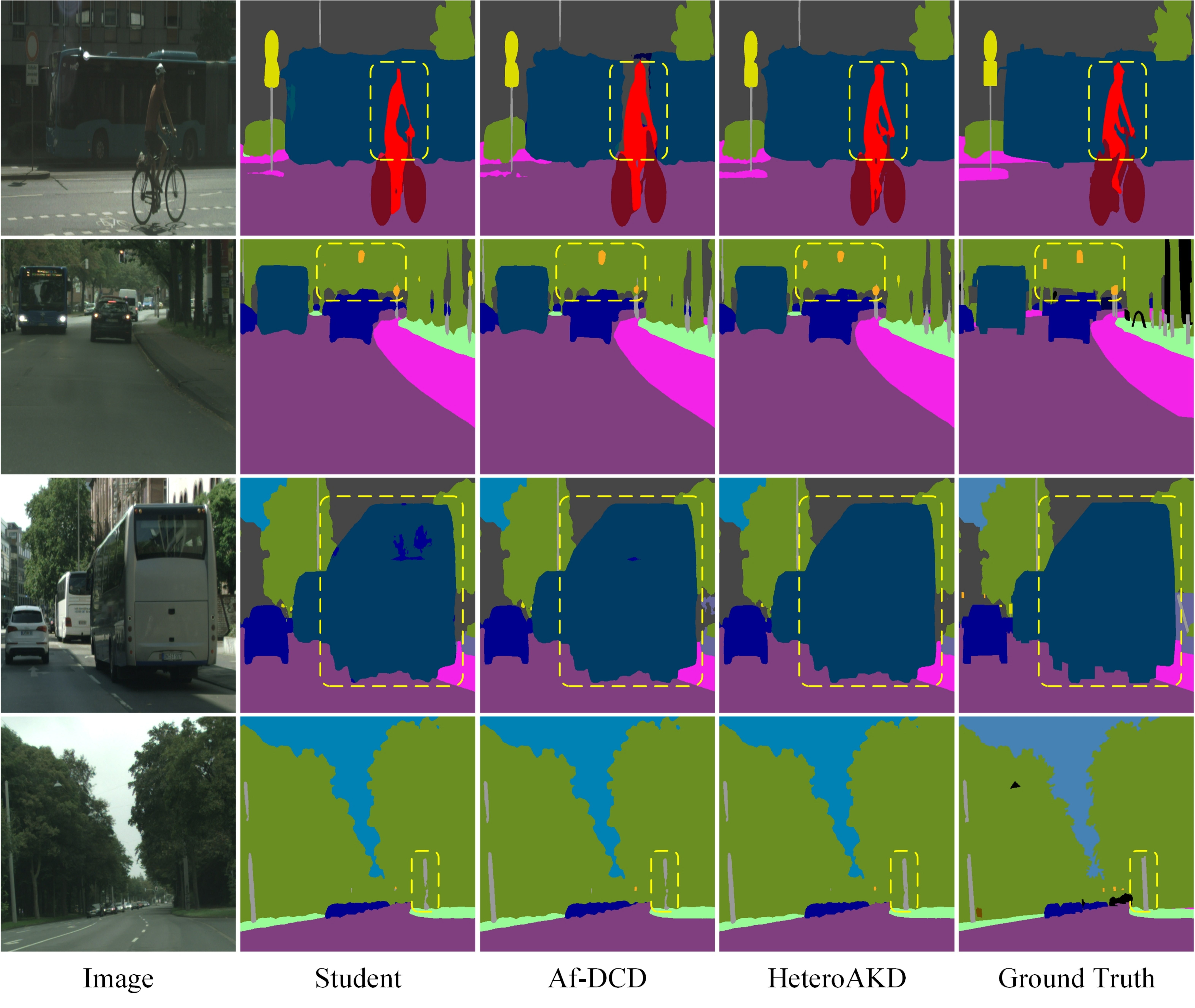}
\caption{Qualitative segmentation results on Cityscapes validation set. The first and second rows are results on SegFormer-MiT-B4$\rightarrow$DeepLabV3-Res18 pair, while the third and fourth rows are results on DeepLabV3-Res101$\rightarrow$SegFormer-MiT-B1 pair.}
\label{fig7}
\end{figure*}

\subsection{Detailed Analysis of Hyper-parameters}

\subsubsection{Impact of the Temperature $\tau$:}
Temperature $\tau$ is used to calibrate the smoothing of the predicted distributions. A more significant temperature $\tau$ brings a smoother distribution. Our method obtains the best results when setting $\tau=1.0$ and $\tau=0.7$ in each of the two modes, which is similar to the empirical settings in CIRKD~\cite{liu2019s,yang2022c}. 

\subsubsection{Impact of the Coefficient $\lambda_{1}$:}
Coefficient $\lambda_{1}$ is employed to control the weight of $\mathcal{L}_{kd}$, which determines the extent to which the student directs their attention to the final output of the teacher during the training process. The experimental results show that $\lambda_{1}=0.1$ is a relative proper setting on both modes.

\subsubsection{Impact of the Coefficient $\lambda_{2}$:}
Coefficient $\lambda_{2}$ is utilized to control the weight of $\mathcal{L}_{hakd}$, which affects the strength of knowledge transfer for intermediate features.
Our method demonstrates a relatively stable performance across varying values of the coefficient $\lambda_{2}$. However, it is worth noting that the optimal coefficient values for the two modes are 1.0 and 10.0, respectively. This difference suggests that different heterogeneous teacher-student pairs exhibit inconsistency in facilitating the learning of intermediate features. We attribute this phenomenon to varying disparities in knowledge between heterogeneous teacher-student pairs. A higher value of the coefficient $\lambda_{2}$ indicates that the student can acquire more knowledge from the heterogeneous teacher.

\section{Visualization}

\subsection{Qualitative Segmentation Visualization}
To intuitively analyze the effectiveness of the proposed \textit{HeteroAKD}, we show the qualitative segmentation results in Figure~\ref{fig7}. We can observe that our \textit{HeteroAKD} produces more consistent semantic labels with the ground truth than baseline and Af-DCD~\cite{fan2023a}. Specifically, our \textit{HeteroAKD} can help student to classify difficult pixels (cases in the first and second rows) and to segment object more completely (cases in the third and fourth rows).

\end{document}